\documentclass{llncs}
\usepackage{makeidx}  
\usepackage{etoolbox} 
\usepackage{hyperref} 
\usepackage{booktabs}
\usepackage{graphicx}
\usepackage{pifont}
\usepackage{xcolor}

\usepackage{hyperref}
\hypersetup{
    colorlinks=true,
    linkcolor=blue,
    filecolor=magenta,      
    urlcolor=magenta,
    pdftitle={Overleaf Example},
    pdfpagemode=FullScreen,
    }
\newcommand\RaiseImage[2][scale=1]{%
  \raisebox{-0.5\totalheight}{\includegraphics[#1]{#2}}}
  
\makeatletter
\patchcmd{\ps@headings}{\rlap{\thepage}}{}{}{}
\patchcmd{\ps@headings}{\llap{\thepage}}{}{}{}
\makeatother
\usepackage{float}
\usepackage{makecell}
\definecolor{cvprblue}{rgb}{0.21,0.49,0.74}
\usepackage{framed}
\usepackage{tikz} 
\usepackage{adjustbox}
\usepackage{multirow}
\usepackage{array}

\definecolor{myblue}{RGB}{218, 235, 253}
\definecolor{mygreen}{RGB}{214, 226, 211}
\definecolor{myyellow}{RGB}{245, 229, 164}
\definecolor{myred}{RGB}{236, 212, 209}
\definecolor{mygray}{RGB}{194, 192, 182}
\definecolor{tablegray}{rgb}{0.2,0.2,0.2}

\usepackage{threeparttable}

\begin{document}
%
%

%
\mainmatter              
%
\title{Synthetic data augmentation for robotic mobility aids to support blind and low vision people}
\titlerunning{Hamiltonian Mechanics}  
%

\author{Hochul Hwang \and Krisha Adhikari \and Satya Shodhaka \and Donghyun Kim}
%
\authorrunning{Hochul Hwang et al.} 
%
%
\institute{University of Massachusetts Amherst, USA}

\maketitle              


\begin{abstract}
Robotic mobility aids for blind and low-vision (BLV) individuals rely heavily on deep learning-based vision models specialized for various navigational tasks. However, the performance of these models is often constrained by the availability and diversity of real-world datasets, which are challenging to collect in sufficient quantities for different tasks. In this study, we investigate the effectiveness of synthetic data, generated using Unreal Engine 4, for training robust vision models for this safety-critical application. Our findings demonstrate that synthetic data can enhance model performance across multiple tasks, showcasing both its potential and its limitations when compared to real-world data. We offer valuable insights into optimizing synthetic data generation for developing robotic mobility aids. Additionally, we publicly release our generated synthetic dataset to support ongoing research in assistive technologies for BLV individuals, available at \url{https://hchlhwang.github.io/SToP}. 

\keywords{Perception, Reasoning and inference, Artificial intelligence}
\end{abstract}
\begin{figure}
    \centering
    \includegraphics[width=\columnwidth]{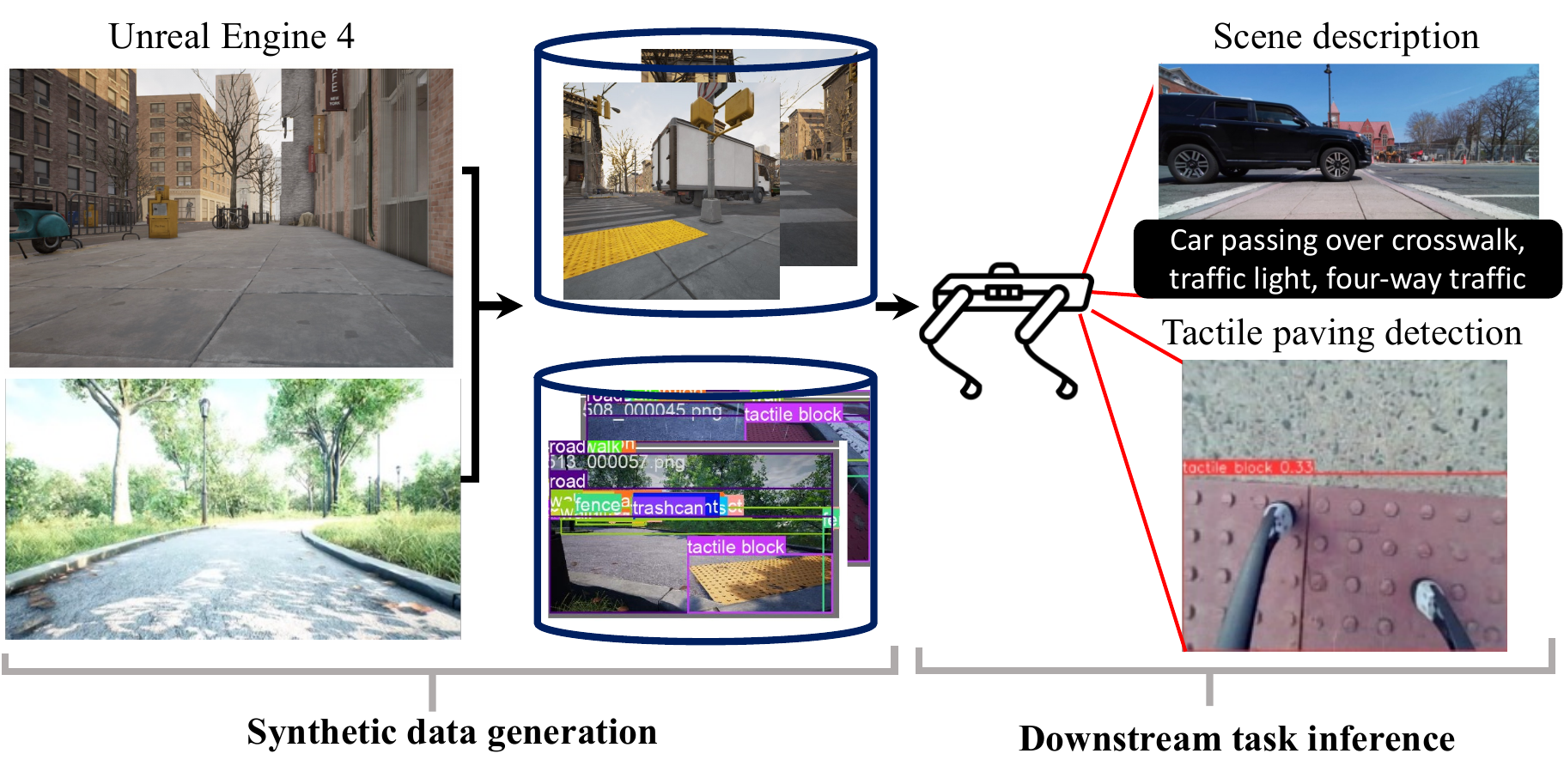}
    \caption{{\bf Synthetic data generation pipeline.} We generated synthetic data using Unreal Engine 4 and the NVIDIA Deep Learning Dataset Synthesizer for various navigational downstream tasks.}
  \label{fig:scheme}
\end{figure}
\section{Introduction}
%

Globally, over 250 million people are blind or have low-vision (BLV), with estimates suggesting that this number could exceed 700 million by 2050~\cite{ackland2017world}. Orientation and mobility are foundational skills that enable BLV individuals to navigate their environments safely and effectively. Orientation involves understanding one's location and direction relative to landmarks and maintaining spatial awareness, while mobility focuses on the ability to move safely and efficiently through an area~\cite{wiener2010foundations}. 

Traditional mobility aids, such as guide dogs and long canes, provide essential support in mobility for BLV individuals to navigate safely around obstacles~\cite{whitmarsh2005benefits}. However, these solutions have limitations: guide dogs are accessible to only a small fraction of the BLV population due to their limited availability and the considerable commitment required for their care, while long canes, although more accessible, impose a significant cognitive burden on users who must constantly detect and avoid obstacles. In response to these challenges, various robotic mobility aids have been developed~\cite{guerreiro2019cabot,xiao2021robotic,hwang2023system}, offering the potential to enhance mobility and independence for BLV individuals.

Robotic mobility aids typically rely on deep learning models to interpret the complex, dynamic environments encountered by BLV users. Training these models requires extensive annotated data, which is challenging to obtain across different systems and may lack the diversity needed for robust performance across different tasks. For example, different robotic aids may operate from different viewpoints, and tasks such as scene description and object detection require distinct data types, further complicating data collection and model training.

Recent advances in computer vision and deep learning have highlighted synthetic data as a viable alternative to real-world data, especially in scenarios where real data is scarce or difficult to obtain. Synthetic data allows for the generation of large-scale datasets with diversity in environmental conditions and scenarios, crucial for training robust models. Moreover, it offers precise control over the complexity and variability of scenes, promoting the development of generalized and adaptable models.

In this work, we present a synthetic data generation pipeline designed to enhance the performance of assistive robotic systems for BLV individuals as shown in Fig.~\ref{fig:scheme}. Our generated synthetic datasets are tailored for key tasks, including tactile paving detection and scene description, which can be critical for safe street crossing. We evaluate the effectiveness of state-of-the-art object detection and vision-language models trained on our synthetic data, demonstrating that synthetic data can improve model performance across these tasks. To facilitate further research and development in assistive technologies, we have made our synthetic dataset publicly available.

The key contributions of our work can be summarized in threefolds: 
\begin{itemize}
    \item We propose a pipeline for generating synthetic data tailored for training deep learning models used in robotic mobility aids. 
    \item We demonstrate the effectiveness of synthetic data in fine-tuning models for downstream tasks, showing improved performance in tactile paving detection and scene description.
    \item We share a comprehensive synthetic dataset that includes a wide range of scenarios, enhancing the robustness of models across various tasks. 
\end{itemize}

\section{Related work}
\subsection{Robotic mobility aids and datasets}
Robotic mobility aids come in various forms—such as smart canes~\cite{slade2021multimodal}, cart-shaped devices~\cite{guerreiro2019cabot,kuribayashi2023pathfinder}, drones~\cite{avila2017dronenavigator,tan2021flying}, and legged robots~\cite{xiao2021robotic,hwang2023system}—and provide physical support beyond orientation aids, which offer only orientation information usually through mobile devices~\cite{yu2019lytnetv2,oko,bme}. These robotic mobility aids rely on visual perception systems to understand their surroundings and ensure safe navigation. While large-scale datasets developed for autonomous vehicles, such as KITTI~\cite{geiger2013vision} and Cityscapes~\cite{cordts2016cityscapes}, have driven advances in deep learning models, their domain-specific characteristics often limit their applicability to sidewalk environments~\cite{zhang2024towards}. For example, a legged guide robot that operates close to the ground on sidewalks requires data distinct from that acquired by vehicles on roads.

With such consideration, recent efforts made strides in providing data related to sidewalk environments for developing efficient mobility aids. The SideGuide dataset~\cite{park2020sideguide} comprises over 490K images annotated with object bounding boxes and polygon segmentation masks. This dataset, partly crowdsourced and collected from wheelchair-mounted ZED cameras, offers a fixed viewpoint from wheelchair users. In contrast, the VIsual Dataset for Visually Impaired Persons (VIDVIP)~\cite{baba2021vidvip} targets applications specifically for BLV individuals, containing images from various Japanese cities and annotations for 39 object classes, including unique sidewalk features such as tactile pavings and signal buttons. Although VIDVIP consists of only 32,000 images, it includes approximately 540,000 annotated instances, addressing sidewalk characteristics not covered by SideGuide. However, some classes are location-specific and may not be relevant in other countries, like the United States. Waghmare et al.~\cite{waghmare2023sanpo} introduced SANPO, a video dataset featuring outdoor environments from a human egocentric viewpoint, with 112K real panoptic masks and 113K synthetic panoptic masks. This dataset provides ground truth labels for instance segmentation and depth estimation, enhancing the utility of synthetic data in training models.

\begin{figure}
    \centering
    \includegraphics[width=\columnwidth]{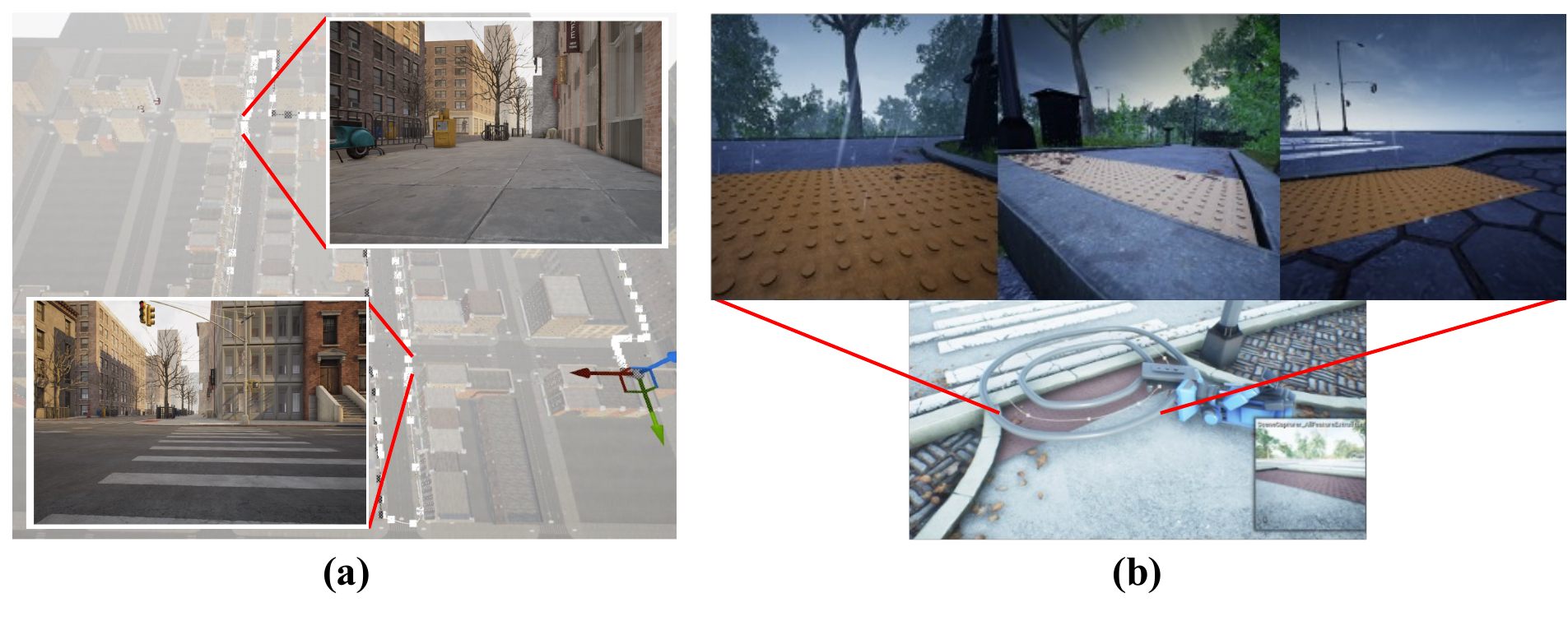}
    \caption{{\bf Synthetic data generation environment.} (a) The Suburban environment features urban roads and sidewalks that contain a variety of objects commonly found in sidewalk settings. (b) Controllable camera trajectories allow data collection from diverse viewpoints, reflecting the perspectives of different robotic mobility aids.}
  \label{fig:background}
\end{figure}

\subsection{Tactile paving detection}
Tactile pavings are designed as textured ground indicators to assist BLV people in navigating sidewalk environments which play a crucial role in BLV people's travel and assist them in finding the way forward. Early methods for tactile paving detection, such as the approach proposed by Ghilardi et al.~\cite{ghilardi2016new}, utilized traditional computer vision algorithms and decision trees to detect tactile pavings in low-resolution images, but required a fixed viewpoint. Ito et al.~\cite{ito2021tactile} demonstrated tactile paving detection by dynamic thresholding based on HSV space analysis for developing a walking support system. As mentioned in their analysis, such methods may be sensitive to viewpoint and illumination changes. 

To achieve more robust tactile paving detection, recent approaches leverage deep learning to learn consistent features. Zhang et al.~\cite{zhang2024grfb} introduced a segmentation algorithm combining UNet~\cite{ronneberger2015u} with a multi-scale feature extractor to capture detailed texture information. Their dataset, collected from the human perspective, reflects the ongoing challenge of acquiring adequate real-world data. In response, our work focuses on generating synthetic data to train deep learning models for tactile paving detection and other downstream tasks, addressing the limitations of existing datasets and enhancing the robustness of assistive technologies for BLV individuals.

\section{Synthetic Data Generation Pipeline}
In this work, we propose a synthetic dataset generation pipeline utilizing Unreal Engine 4 (UE4) with the NVIDIA Deep Learning Dataset Synthesizer (NDDS) plugin~\cite{to2018ndds} for photorealistic rendering and automated data annotation. This approach enables the generation of ground-truth labels and synthetic data customized for specific robotic mobility aids, accounting for variations in viewpoints and lighting conditions. The synthetic data generation environment of the pipeline is illustrated in Fig.~\ref{fig:background}.


\subsection{Background}
We collected data from two Environment packages available from the UE4 Marketplace, designed to reflect real-world scenarios. The City Park environment covers land with varying terrains, objects, and road paths, closely resembling a typical park setting. This pre-made environment includes approximately 800 objects commonly found in parks, such as benches, water fountains, and varying vegetation. The Suburban environment includes urban roads and sidewalks populated with a variety of objects such as buildings, traffic lights, and curbs (see Fig.~\ref{fig:background} (a)). Although this environment is smaller in scale, it is densely packed with around 2,000 objects, providing a detailed urban setting compared to the city park. To enhance the realism of the synthetic data, we generated datasets under varying lighting and weather conditions, reflecting the complexities of real-world environments. 

\subsection{Viewpoint}
Our pipeline leverages UE4's flexibility to simulate various viewpoints corresponding to different robotic mobility aids. This allows for efficient collection of synthetic data from multiple perspectives, accommodating the unique camera placements of these aids (e.g., front, side, and bottom-facing cameras), which often present significant differences in visual appearance. In both environments, we established a range of camera trajectories to capture diverse viewpoints, ensuring a comprehensive representation of the perspectives likely encountered by robotic systems in real-world applications as shown in Fig.~\ref{fig:background} (b).

\subsection{Object}
In addition to utilizing pre-built objects within the selected environments, we created task-specific objects to meet the unique needs of our applications. For instance, to support the development of models for tactile paving detection, we designed custom tactile paving objects not originally present in the environments. We enhanced the realism of these objects by applying high-quality textures sourced from material packs created by third-party developers, ensuring that the synthetic data closely resembles the visual characteristics of real-world objects and materials.

\section{Synthetic Data}
We introduce a specialized synthetic dataset generated using our proposed pipeline, tailored for specific tasks such as tactile paving detection and scene description. This dataset is publicly available to support further research and development in assistive technologies for BLV individuals.

\begin{figure}
    \centering
    \includegraphics[width=\columnwidth]{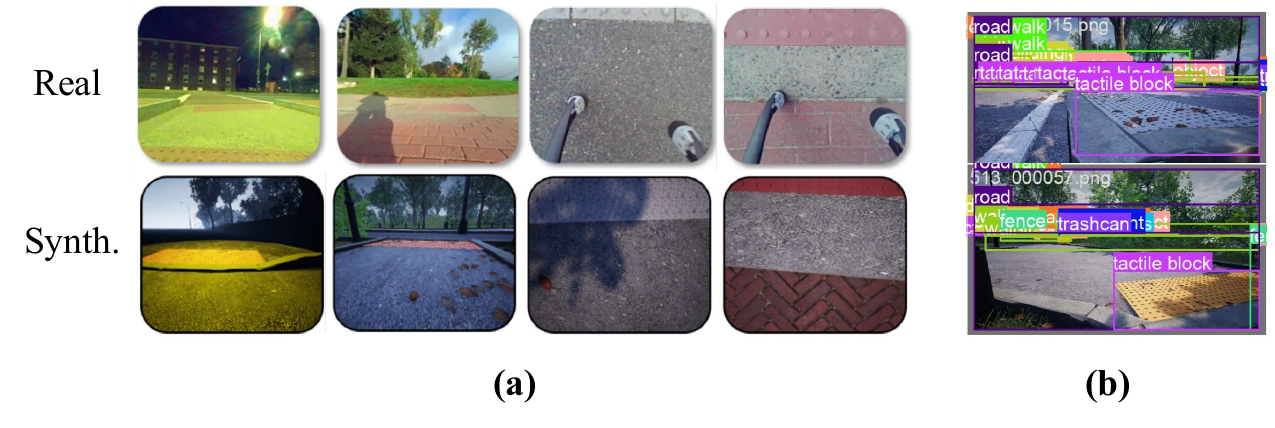}
    \caption{{\bf Samples from SToP Dataset.} (a) Comparison between real data and generated synthetic data in various lighting and viewpoint settings, highlighting the close resemblance of synthetic data to real-world conditions. (b) Visualization of ground truth bounding boxes within the UE4 environment.}
  \label{fig:stop}
\end{figure}


\subsection{SToP Dataset}
The Synthetic Tactile-on-Paving (SToP) Dataset is designed to enhance the performance of perception systems in object detection and semantic segmentation tasks, specifically targeting the detection of tactile pavings essential for navigation for BLV people. Utilizing the NDDS toolset, the dataset includes comprehensive features such as object bounding boxes, segmentation masks, depth information, and camera intrinsics. The tactile paving objects in the dataset were meticulously designed based on the guidelines provided by the American Disability Association (ADA)~\cite{ada}, with dimensions aligned to ADA braille measurement standards to create blister-type tactile blocks, predominantly found in the United States. We employed the Material Pack Tactile Blocks~\cite{tactile} to apply realistic textures, ensuring that the blocks varied in color and appearance to reflect real-world diversity.

To accommodate different viewpoints, we implemented three distinct camera trajectories for data collection in both the City Park and Suburban environments: a wide-circular angle, a circular angle, and a top-down view. This approach ensures that the dataset captures a broad spectrum of perspectives, enhancing the robustness of models trained on this data for real-world applications as shown in Fig.~\ref{fig:stop}.

\begin{figure}
    \centering
    \includegraphics[width=\columnwidth]{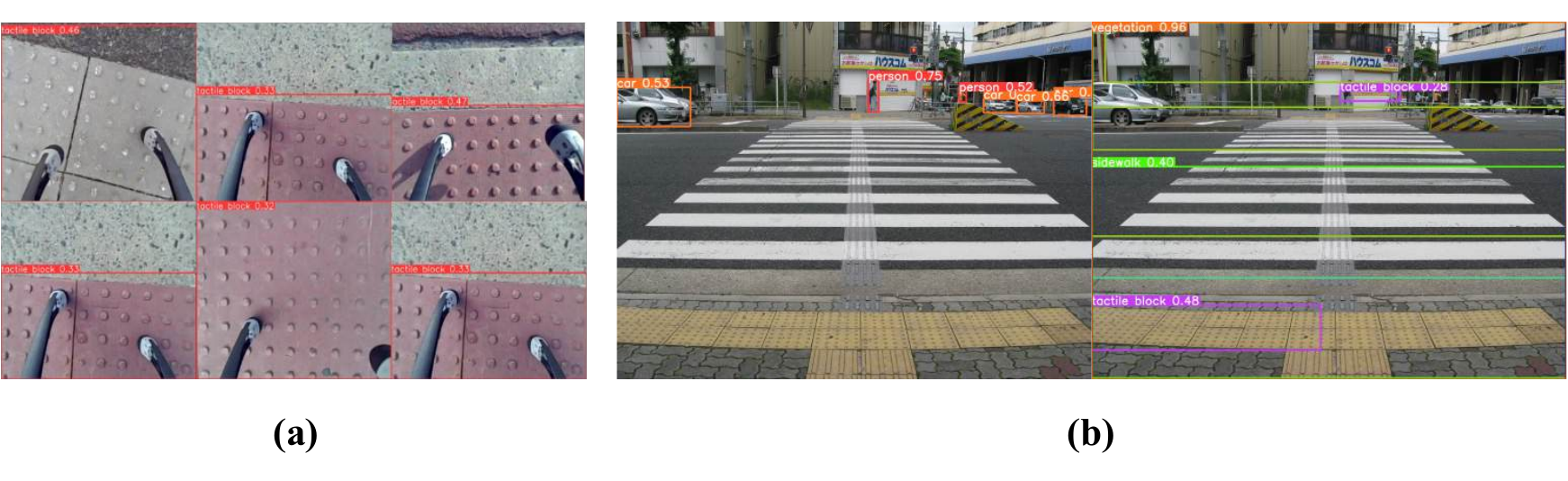}
    \caption{{\bf Tactile paving detection results.} (a) YOLOv8 successfully detects tactile pavings from a top-down view, which were not detected by the pretrained model without synthetic data training. (b) The open-vocabulary YOLO-World provides bounding boxes for tactile pavings (right), a capability that was not achieved previously (left) on a publicly available dataset~\cite{yu2019lytnetv2}.}
  \label{fig:preds}
\end{figure}

\subsection{Synthetic Street Crossing Dataset}
The Synthetic Street Crossing Dataset is developed to fine-tune foundation models for scene description tasks, particularly focused on crosswalk scenarios. Following approaches similar to prior work~\cite{hwang2024safe}, we populated the environment with various vehicles (e.g., trucks and cars) and pedestrian signals to accurately reflect real-world street crossing situations. One of our researchers manually annotated the dataset, providing scene descriptions in text that align with the preferences and needs of BLV individuals.

\section{Experimental Result}

\subsection{Tactile paving detection}
To evaluate the SToP dataset, we fine-tuned YOLOv8~\cite{Jocher_Ultralytics_YOLO_2023} and YOLO-World~\cite{cheng2024yolo} specifically for tactile paving detection. We used 3,000 image-bounding box pairs from the SToP dataset to train the YOLOv8m and YOLO-World models for 190 epochs, with a learning rate set to 0.01. We present qualitative results, demonstrating the model’s predictions on synthetic data as well as on real-world data collected using a Unitree Go1~\cite{Unitree} quadruped robot. The real-world dataset features a top-down viewpoint, captured by the robot’s camera facing the ground, closely mirroring the viewpoint expected in practical applications.

The result in Fig.~\ref{fig:preds} shows that YOLOv8 and YOLO-World trained on the SToP dataset effectively detect tactile pavings in both synthetic and real-world scenarios. This indicates that synthetic data, when generated with attention to realistic textures and viewpoints, can enhance the detection of target objects. 


\subsection{Scene description}
For the scene description task, we fine-tuned the Florence-2~\cite{xiao2024florence} vision foundation model using our synthetic street crossing dataset, complemented by text annotations crafted by one of the researchers to reflect the informational preferences of BLV individuals during street crossings. The annotations were designed to include critical details such as the presence of vehicles, pedestrian signals, and crossing statuses that are most relevant to BLV users.

\begin{table}[t]
    \centering
    \begin{threeparttable}
    \begin{tabular}{c|cccc}
        \toprule 
        \textbf{Method} & \textbf{Precision}& \textbf{Recall} & \textbf{F1} & \textbf{BLEU}  \\
        \hline\hline
        Baseline & 0.9273 & 0.8996  & 0.9130 & 0.2419\\
        \hline
        + Real & \textbf{0.9278} &\textbf{0.9074}  & \textbf{0.9173}  & 0.2635 \\
        + Synth. & 0.9248 & 0.9049  & 0.9144  & \textbf{0.2689} \\
        \bottomrule
    \end{tabular}
    \caption{\textbf{Scene description results.} }
    \label{tab:scenequant}


    \end{threeparttable}
\end{table}

\begin{table*}[ht]
\centering
\begin{tabular}{m{1cm}|m{5.5cm}|m{5.5cm}}
\toprule

\textbf{Img.} & \RaiseImage[width=4.5cm, height=4.5cm]{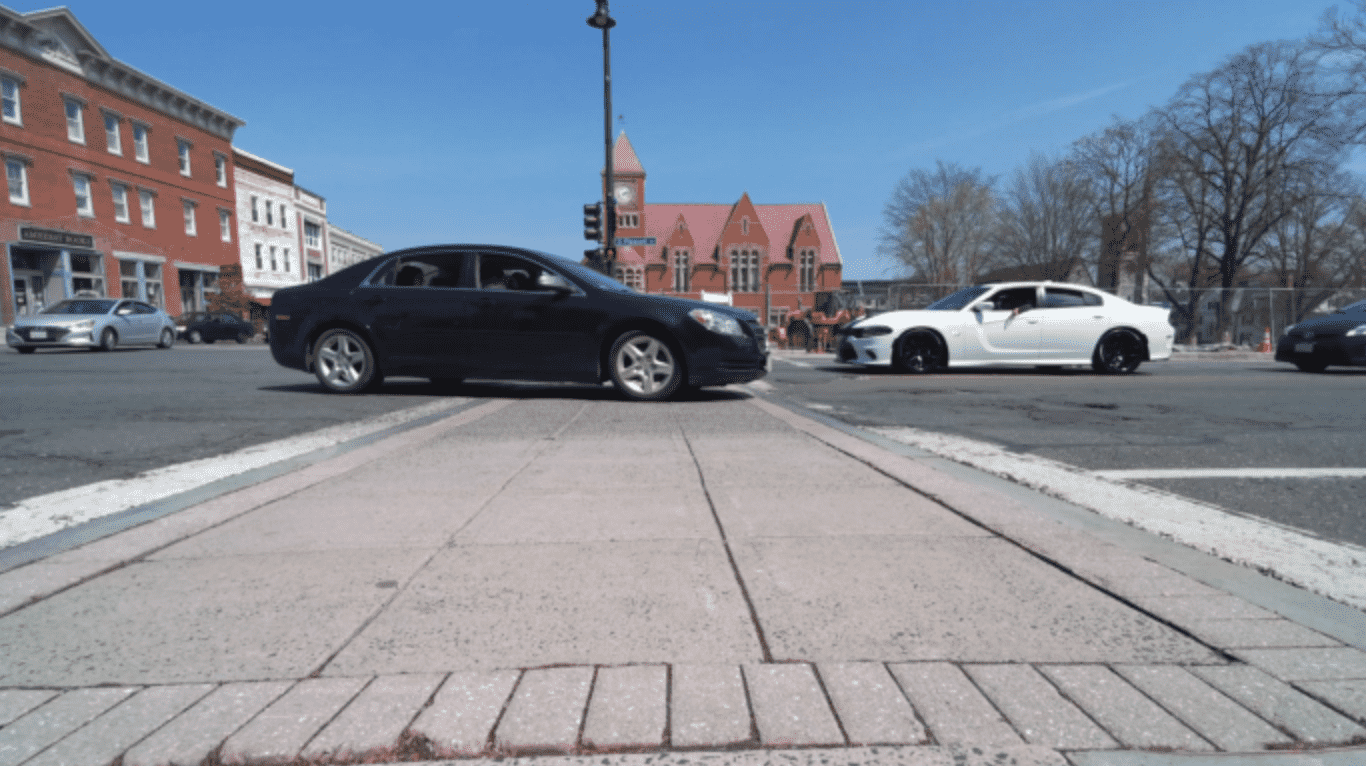} & \RaiseImage[width=4.5cm, height=4.5cm]{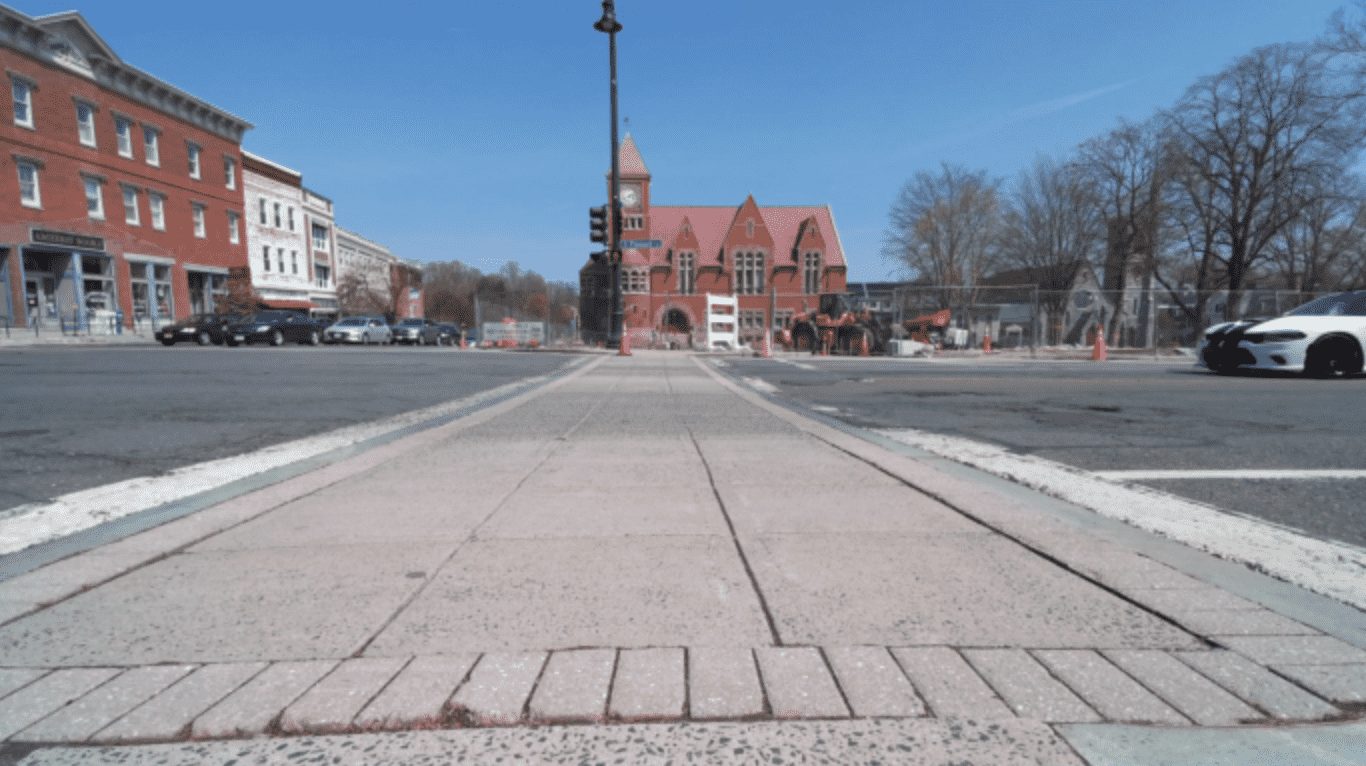}
\\ \midrule
\textbf{GT.} &car passing over crosswalk from the left side, pedestrian light red, four-way traffic & pedestrian light red, car approaching crosswalk from right, four-way traffic
\\ 
\midrule
\textbf{Pred.} & car passing over crosswalk from left to right, four-way traffic & pedestrian light red, four-way traffic
\\
\bottomrule
\end{tabular}
\caption{\textbf{Qualitative results for the scene description task.} } 
\label{tab:scenedesc}
\end{table*}



We evaluated the fine-tuned model on a test dataset consisting of 400 samples collected using an Azure Kinect camera mounted on a Go1 quadruped robot. Quantitative analysis of the scene description task included metrics such as BLEU scores, precision, and recall. For metrics, BERT~\cite{devlin2018bert} embeddings were used to represent the words as tokens and cosine similarity was used to select the most similar token from the reference to calculate the scores. 

As shown in Table~\ref{tab:scenequant}, incorporating additional real-world and synthetic data led to comparable performance improvements over the baseline, except for precision. Although both types of data augmentation yielded similar benefits, real-world data demonstrated slightly higher performance in terms of precision and recall. The higher BLEU scores highlight the potential of synthetic data in enhancing vision-language models to generate the precise wording preferred in scene descriptions. 

Qualitative results of the fine-tuned Florence-2 model are presented in Table~\ref{tab:scenedesc}, where the model exhibited strong performance in generating accurate descriptions across varied conditions, though some inaccuracies and missing information were found. For example, the model did not provide information on the pedestrian light in the first example in Table~\ref{tab:scenedesc}.







\section{Conclusion}

In this paper, we introduced a synthetic data generation pipeline leveraging a game engine to train deep learning models for robotic mobility aid applications. Our approach focused on two primary tasks: tactile paving detection and scene description. By fine-tuning the YOLOv8 and YOLO-World models on our synthetic dataset, we successfully enhanced the detection of tactile pavings, addressing gaps that previous pretrained models had struggled with. Additionally, we fine-tuned the Florence-2 vision foundation model to generate scene descriptions that align more closely with the preferences of BLV individuals. 

While our results highlight the potential of synthetic data, we acknowledge the existing gap between synthetic and real-world data, as well as the complexities involved in creating realistic virtual environments, which require specialized expertise in game engine use. Moreover, our work is limited by the relatively small amount of data (seven additional images) used for scene description, and challenges remain in scenarios with extreme lighting variations or occlusions for tactile paving detection. 

Despite these challenges, synthetic data generation potentially offers a more efficient alternative to collecting and annotating real-world data, providing the flexibility to tailor environments and datasets to specific application needs. This adaptability makes synthetic data a valuable resource for advancing assistive technologies, particularly in scenarios where real-world data acquisition is impractical or resource-intensive. 





\section*{Acknowledgement}
We would like to thank Shiven Patel, Antoinette Reid, Dang Nguyen, and Ron Kleinhause-Goldman from the CICS Early Research Scholars Program at the University of Massachusetts Amherst for their valuable assistance in collecting the test data for the scene description task.

%
%
\bibliographystyle{plain}  
\bibliography{reference}







\end{document}